\documentclass[runningheads]{llncs}
\usepackage{graphicx}
\usepackage{caption}
\usepackage{subcaption}
\usepackage{soul}
\usepackage{color}
\usepackage{xcolor}
\usepackage{javaLatexSTY/lstcustom}

\definecolor{pblue}{rgb}{0.13,0.13,1}
\definecolor{pgreen}{rgb}{0,0.5,0}
\definecolor{pred}{rgb}{0.9,0,0}
\definecolor{pgrey}{rgb}{0.46,0.45,0.48}

\newcommand{\hcode}[1] {\textbf{\textcolor{purple}{\texttt{#1}}}}

\begin{document}
\title{Towards High Performance Java-based Deep Learning Frameworks}
%
%
\author{Athanasios Stratikopoulos\inst{1} \and Juan Fumero\inst{1} \and Zoran Sevarac\inst{2} \and Christos Kotselidis\inst{1}}
%
\authorrunning{A. Stratikopoulos, J. Fumero, Z. Sevarac and C. Kotselidis}
%
\institute{The University of Manchester, Manchester, United Kingdom \\
\email{\{first.last\}@manchester.ac.uk} \and
Deep Netts LLC, Belgrade, Serbia \\
\email{zoran.sevarac@deepnetts.com}}

\maketitle              
\begin{abstract} 
The advent of modern cloud services along with the huge volume of data produced on a daily basis, have set the demand for fast and efficient data processing. 
This demand is common among numerous application domains, such as deep learning, data mining, and computer vision. 
Prior research has focused on employing hardware accelerators as a means to overcome this inefficiency. 
This trend has driven software development to target heterogeneous execution, and several modern computing systems have incorporated a mixture of diverse computing components, including GPUs and FPGAs. 
However, the specialization of the applications' code for heterogeneous execution is not a trivial task, as it requires developers to have hardware expertise in order to obtain high performance.
The vast majority of the existing deep learning frameworks that support heterogeneous acceleration, rely on the implementation of wrapper calls from a high-level programming language to a low-level accelerator backend, such as OpenCL, CUDA or HLS. 

In this paper we have employed TornadoVM, a state-of-the-art heterogeneous programming framework 
to transparently accelerate Deep Netts; a Java-based deep learning framework. 
Our initial results demonstrate up to 8x performance speedup when executing the back propagation process of the network's training on AMD GPUs against the sequential execution of the original Deep Netts framework. 

\end{abstract}

\section{Introduction}
In recent years artificial intelligence is gaining more and more popularity, with its main objective being to enable computers to make decisions that are normally made by domain experts. 
This domain is a superset of Machine Learning and Deep Learning, which both rely on training a mathematical model to perform various tasks using historical data related to the application of interest.
Deep learning is a field that emerged recently and uses the structure of an artificial neural network to train a model to autonomously perform a task, such as computer vision, pattern recognition, speech recognition, etc~\cite{Patterson:2017:DLP:3169957}. 

\begin{figure}[t]
	\centering
	\includegraphics[scale=0.6]{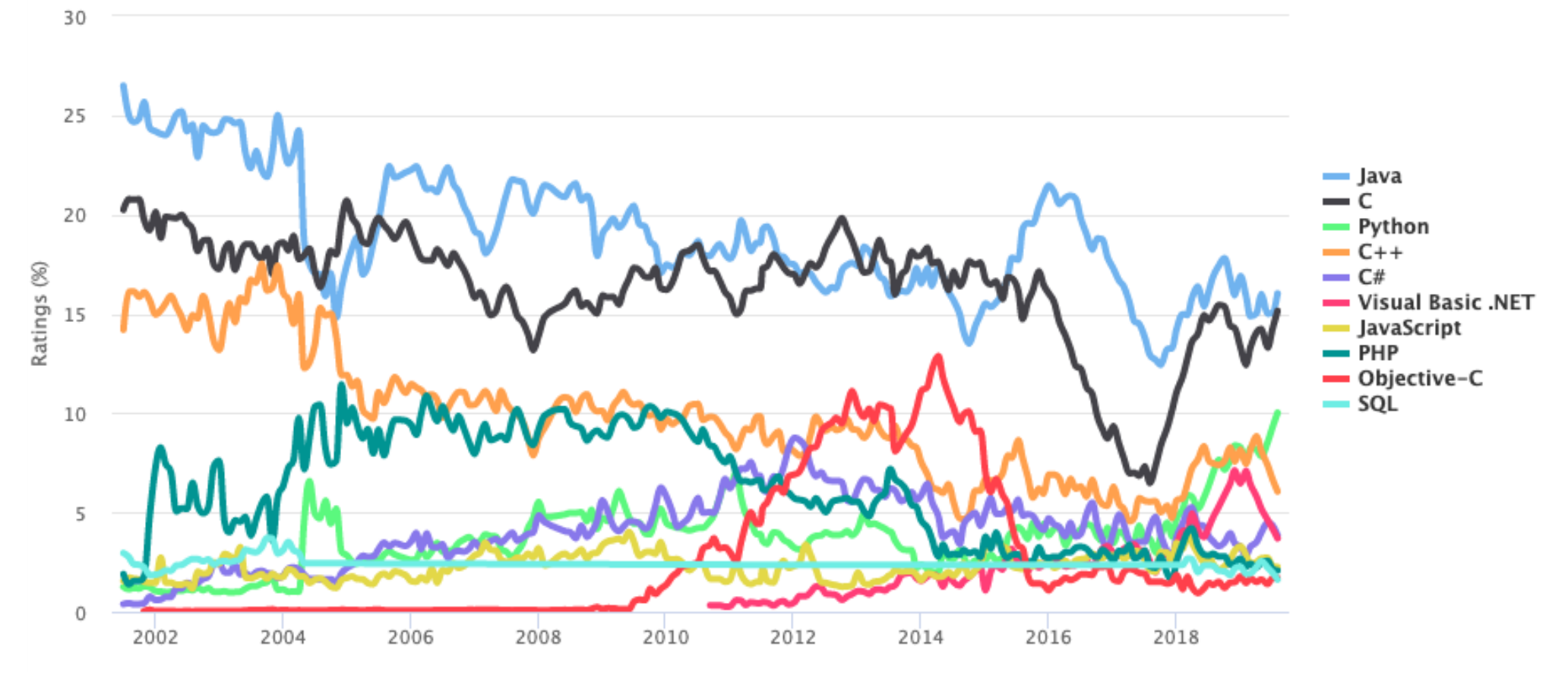}
	\caption{Trend on programming languages. Source: www.tiobe.com.}
	\label{label::trend_languages}
	\vspace{-2em}
\end{figure}

To program machine learning and deep learning applications, developers tend to use high-level abstractions and interpreted programming languages, such as Java, Python, R and Ruby. This is due to the fact that applications written in the aforementioned programming languages are much easier to understand and maintain.
Figure~\ref{label::trend_languages} presents a ranking graph with all programming languages based on the popularity in the ``TIOBE Index for May 2019''. 
As shown in Figure~\ref{label::trend_languages}, Java has been the most popular language among all programming languages since 2002; with a small interpolation with the C language during the period 2012-2015. 
However, the mainstream of the above-mentioned programming languages has been compiled for CPU execution. 
To cope with the advent of heterogeneous systems, multiple programming frameworks have been introduced.  
The vast majority of them supports C and C++ programming languages and compiles them for: a) GPU execution via CUDA or OpenCL, and b) FPGA execution via High Level Synthesis (HLS), and lately OpenCL. 
The execution of Java programs (and managed languages in general) on heterogeneous accelerators is an active research topic with many challenges associated with the compilation, the memory management and the runtime system, which all lie on the core of the managed languages. 

Current approaches such as~TensorFlow~\cite{199317} for Python or Deeplearning4j~\cite{deeplearning4j} for Java rely on existing pre-compiled code that is optimized for the target architecture. 
However, this approach requires the user to familiarize with a new library, and be attached to a particular accelerator, such as a GPU. 
A different approach could be to automatically accelerate various parts of the deep learning frameworks by transparently using all the available hardware resources in contemporary computers, such as a GPU, an FPGA and a multi-core CPU. 
In this case, a runtime system (e.g., the Java Virtual Machine -JVM-) can make use of the hardware resources without the developers' intervention, in order to facilitate applications to comply with the traditional philosophy of Java: \emph{write once, run everywhere} from a heterogeneous perspective.
The state-of-the-art frameworks for automatically accelerating Java code on heterogeneous hardware are Aparapi~\cite{aparapi_citation}, JaBEE~\cite{Zaremba:2012:JFO:2159430.2159439}, IBM GPU J9~\cite{7429325} and TornadoVM~\cite{Fumero:DARHH:VEE:2019}.


In this paper we show our work in progress towards mitigating the gap in programmability, while enabling users to accelerate the deep learning frameworks directly from Java.
Therefore, we have employed TornadoVM~\cite{Fumero:DARHH:VEE:2019} to accelerate a Java-based deep learning engine such as Deep Netts~\cite{sevarac}. 
In detail this paper makes the following contributions:
\begin{itemize}
	\item It overviews the Deep Netts work-flow in order to identify the main parts that can merit by hardware acceleration.
	\item It presents the modifications in the official source code to integrate the TornadoVM API so as to parallelize parts of the vanilla Deep Netts source code.
	\item It evaluates the performance of original Deep Netts against the proposed Tornado-Deepnetts implementation, showing performance speedups up to 2.63x and 8x for small and large datasets, respectively.
\end{itemize}
\section{Deep Learning Overview}

Deep learning relies on an \texttt{artificial neural network (ANN)} which comprises multiple layers of artificial neurons. This structure is first trained using some application-dependent data and then uses the learned model to make intelligent decisions or accurate predictions. 
The main core in an ANN is the artificial neuron historically known as \texttt{perceptron}. An artificial neuron performs three main tasks: 
\begin{itemize}
    \item It accepts some input values either from the data or from the outputs of neurons, depending on the layer in which this neuron belongs to.
    \item It applies some weights and biases to produce an intermediate value known as the \texttt{net input} value. These parameters are learned during the training of the network so as to increase the accuracy.
    \item The \textit{net input} value is forwarded in the \texttt{activation function}, which performs a non-linear transformation.
\end{itemize}

\begin{figure}[!b]
	\vspace{-1em}
	\centering
	\includegraphics[scale=0.4]{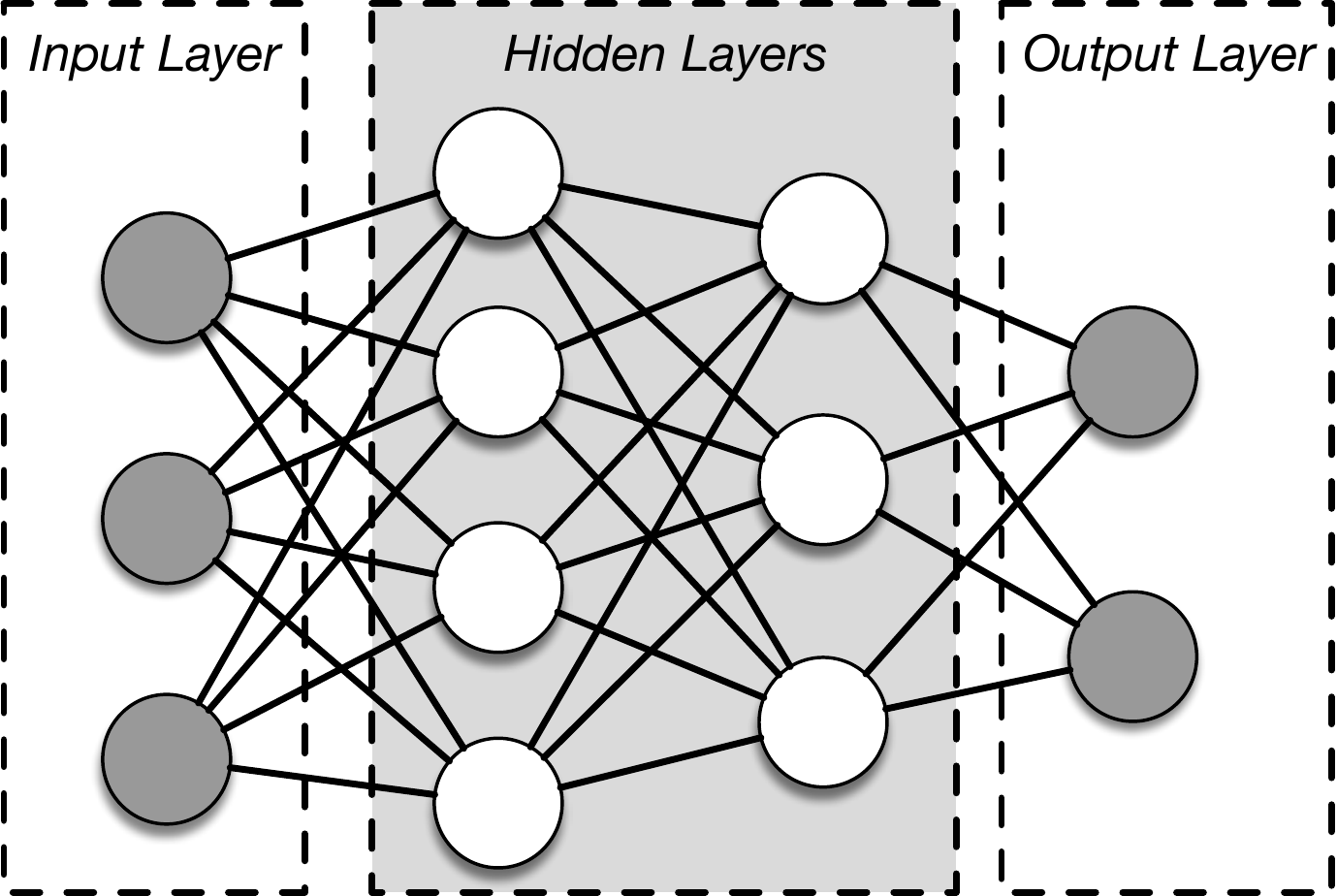}
	\caption{An artificial neural network with fully connected perceptrons.}
	\label{label::ann_fully_connected}
	\vspace{-1.5em}
\end{figure}

The most widely used model of ANNs consists of multiple artificial neurons grouped in three types of layers: \textit{input layer}, the \textit{hidden layers}, and the \textit{output layer}, as shown in Figure~\ref{label::ann_fully_connected}. 
The input layer accepts as input the features from the dataset, and each neuron in this layer can be connected with one or more neurons in the next hidden layer, the outputs of which are connected with the next hidden layer and so on, until the connections reach the output layer. Then, the output layer uses a cost/loss function as a metric to assess the accuracy of the model. 
If the accuracy is not satisfactory, then the back propagation process is performed to update the parameters in each layer so that the average loss decreases. 
This ANN is also called as \texttt{feed forward neural network} as the flow of the information beyond the layers of the network flows forward. 
Other types of neural networks are the \texttt{convolutional} and the \texttt{recurrent} neural networks.
For the rest of this paper we will focus on the feed forward neural network model, which is applicable to a wide range of domains.

\subsection{Deep Netts Deep Learning Engine}
Deep Netts~\cite{sevarac} is a deep learning development platform that focuses on deep learning development in Java. It provides tools for Java developers such as a deep learning library and an integrated development tool, and it enables Java developers to apply deep learning in their applications more easily.
Currently, Deep Netts supports dominant supervised learning algorithms, including linear regression, logistic regression and feed-forward neural networks.
The supported types of neural network layers are: \textit{fully connected} layer, \textit{convolutional} layer, \textit{maximum pooling} layer and the \textit{softmax output} layer.
The key features and advantages of Deep Netts include:
\begin{itemize}
	\item Ease of use, thanks to beginner and Java developer friendly API.
	\item The integration of the state-of-the-art models of neural networks, such as the feed forward network and the convolutional network. These types of networks are provided out-of-box and require less understanding of background theory and library internals in order to be effectively used.
	\item The provision of advanced visualization and logging for understanding, debugging and solving data-based, architecture-based, or training-based, issues~\cite{sevarac}.
	\item Portability, ease of integration, distribution and maintenance (thanks to pure Java implementation), which are features of great importance for large scale deployments.
	\item Deep Netts is a base for reference implementation of standard Java API for visual recognition and machine learning JSR 381~\cite{jsr381}, which is being developed within the official Java technology standardization organization.
\end{itemize}

Nonetheless, one of the main disadvantages of Deep Netts compared to other libraries, is that it lacks support of GPU/FPGA acceleration, so training with large amounts of images, or big images can take a long time. However, the pure Java implementation, along with the clean design and readable code make it suitable for experiments and evolving Java platform towards better support for deep learning and machine learning in general.

\begin{figure}[t]
	\centering
	\includegraphics[scale=0.6]{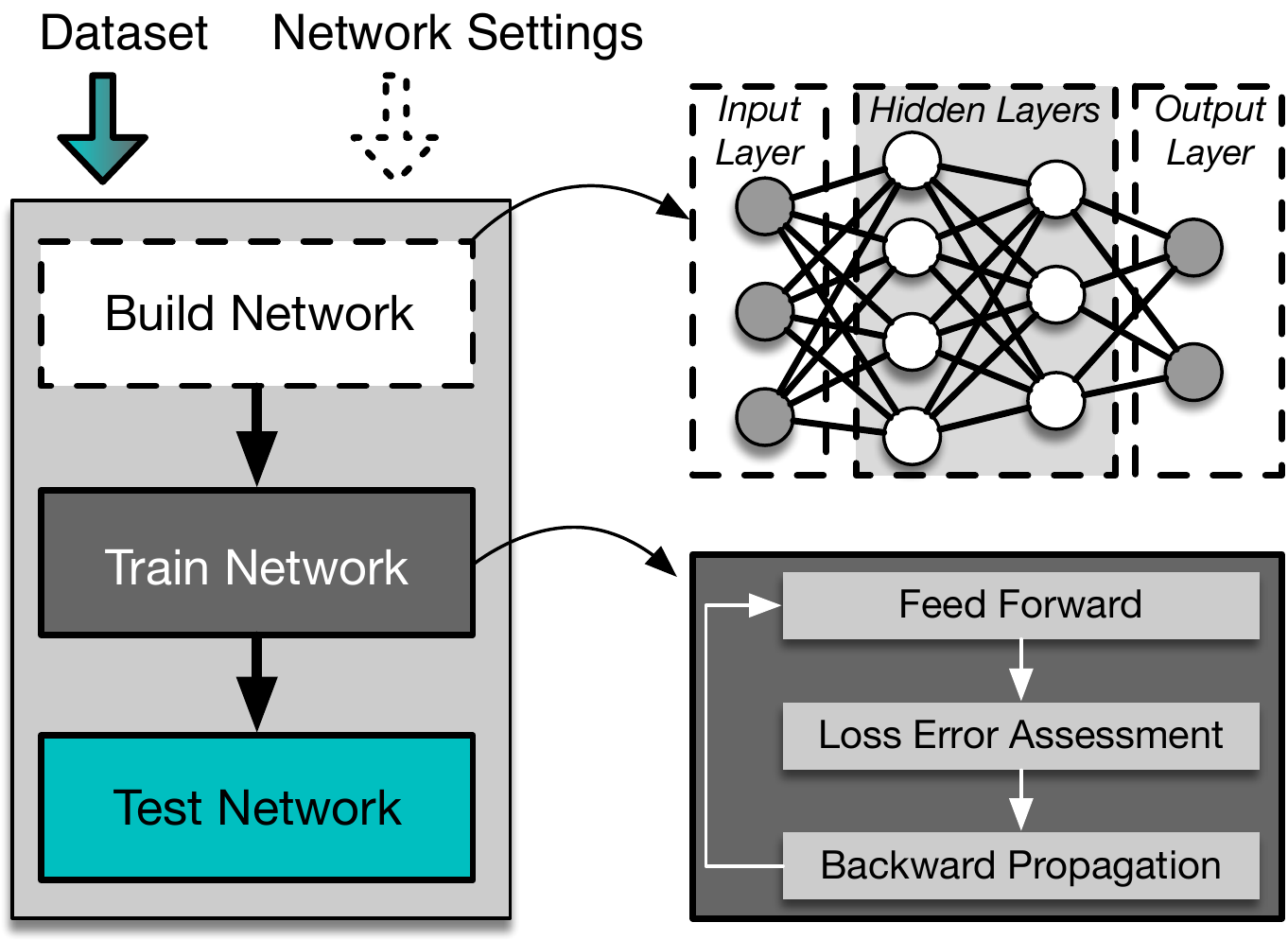}
	\caption{The work-flow of Artificial Neural Networks in Deep Netts.}
	\label{label::deepnetts_flow}
	\vspace{-1.5em}
\end{figure}

\subsubsection{The Deep Netts Work-flow}
Figure~\ref{label::deepnetts_flow} illustrates the main work-flow in Deep Netts framework for building neural networks, and training them to run various algorithms, such as linear regression, neural networks, and logistic regression. The current neural networks in Deep Netts include four types of hidden layers apart from the input and output layers: the \textit{fully connected} layer, the \textit{convolutional} layer, the \textit{maximum pooling} layer and the \textit{softmax} layer.

\begin{figure}[t]
	\centering
	\includegraphics[scale=0.6]{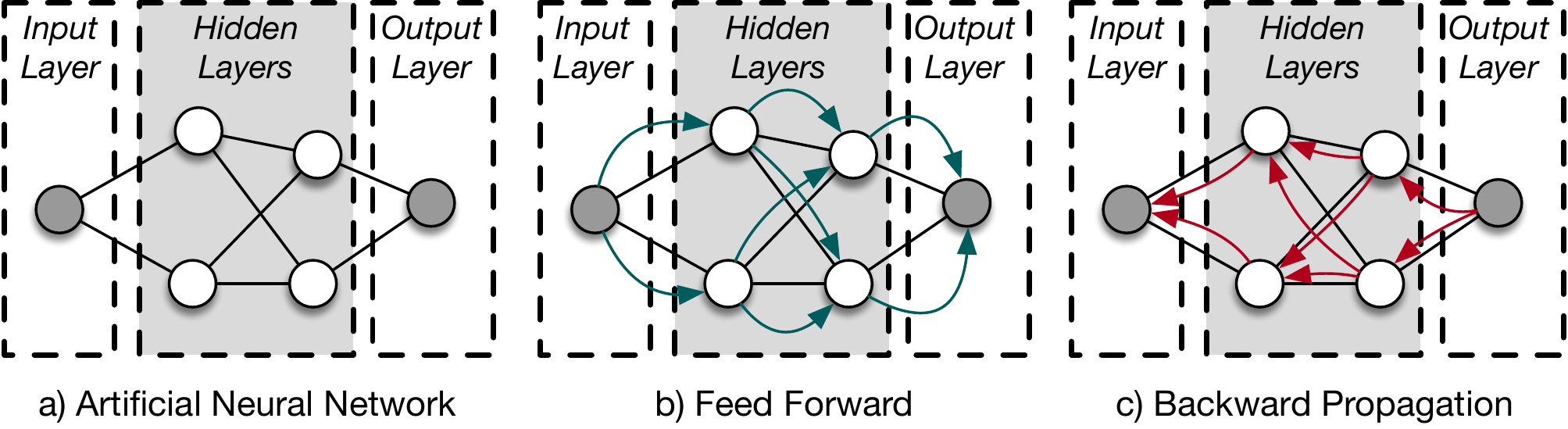}
	\caption{The primary steps of training in Deep Netts. The left figure (a) depicts a fully connected artifical neural network that contains two hidden layers, each of which encloses two neurons. The middle figure (b) shows the feed forward process which includes all activations values from each neuron in the network. The right figure (c) performs the backward propagation that adjusts the configuration of each neuron in order to obtain a highly effective training.}
	\label{label::training_flow}
	\vspace{-1.5em}
\end{figure}

As shown in Figure~\ref{label::deepnetts_flow}, Deep Netts accepts as input the data set upon which the training will be based along with various configuration parameters for the network. These parameters can be the type and the number of layers, and the maximum error rate that the algorithm can tolerate.
Once the inputs are set, the next step is to build the artificial neural networks and initialize the values of weights and biases in each neuron of every layer. Figure~\ref{label::training_flow}a presents a simplistic view of a fully connected built neural network. The next process in the workflow is the training which comprises three parts: 
the feed forward, the loss error assessment, and the back propagation.
\begin{enumerate}
    \item The \textbf{feed forward} triggers the activation functions of each neuron in each layer and creates an activation value which is forwarded to the next layer, until it reaches the output layer. This process is presented in Figure~\ref{label::training_flow}b with the green arrows. The complexity of the feed forward process can be significant, as it is in accordance to the structure of the neural network.
    \item The \textbf{loss error assessment} checks the emerged output from the previous part and makes a comparison with the configured maximum error value. In case the network has not met the accuracy that is required, the back propagation step is performed.
    \item The \textbf{back propagation} traverses all the layers in the reversed order going backwards. This process is responsible for updating the weights and biases in each neuron of the layers in order to increase the overall performance of the model. Figure~\ref{label::training_flow}c represents this process with arrows depicted in red. This part of the training is considered as computationally expensive and can merit to be executed in parallel~\cite{Patterson:2017:DLP:3169957}. Section ~\ref{sec_implementation} will discuss in detail how we parallelized this step for two types of layers: the \textit{fully connected} layer and the \textit{softmax output} layer.
\end{enumerate}

\section{TornadoVM Programming Framework}
\label{section::tornadovm}
TornadoVM is defined as a plugin to OpenJDK that allows Java programmers to automatically execute their applications on heterogeneous hardware. 
Currently, TornadoVM can run on multi-core CPUs, GPUs and FPGAs. 
Additionally, TornadoVM can migrate, at runtime, execution from one device to another~\cite{Fumero:DARHH:VEE:2019} (e.g., from a multi-core CPU to a GPU).

\begin{figure}[t]
	\centering
	\includegraphics[scale=0.4]{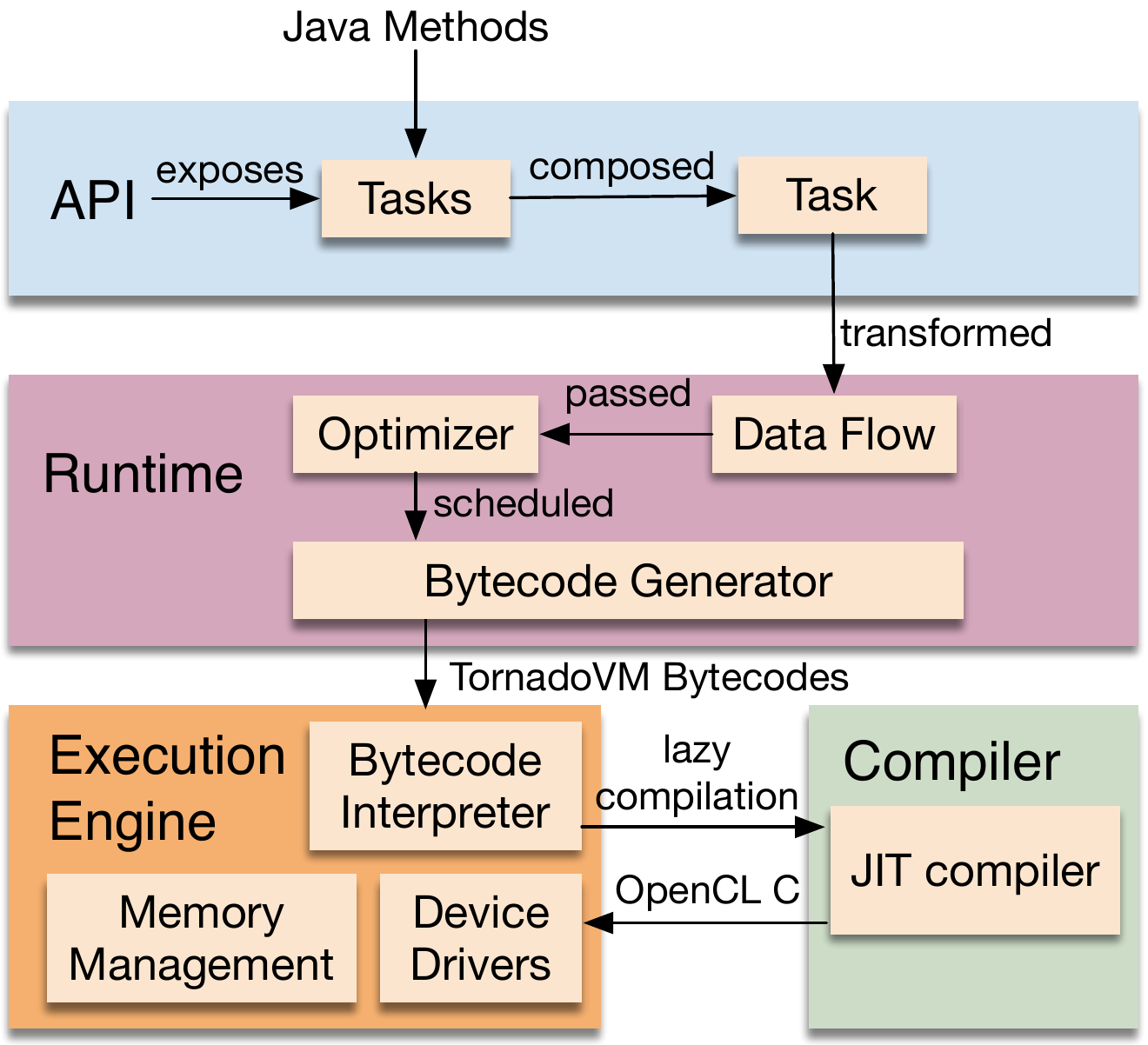}
	\caption{The TornadoVM work-flow.}
	\label{label::tornado_workflow}
	\vspace{-2em}
\end{figure}

Figure~\ref{label::tornado_workflow} shows the three main TornadoVM components (i.e., API, Runtime, Compiler), along with the execution engine which is responsible for the Just In Time (JIT) compilation of the bytecodes and the memory management.
TornadoVM exposes a lightweight API that developers use to indicate which Java methods they would like TornadoVM to accelerate on heterogeneous devices.
Once the user identifies the methods, TornadoVM compiles, at run-time, Java bytecodes to OpenCL C as follows:
a) first, TornadoVM builds a data-flow graph with the aim to optimize the data dependencies between tasks, and subsequently reduce the required data transfers and buffer allocation time;
b) second, TornadoVM generates new bytecodes (TornadoVM Bytecodes) on top of the Java bytecodes, which are used for pure orchestration on the heterogeneous devices; 
c) finally, TornadoVM executes the whole application by using the TornadoVM bytecode, and it compiles at runtime, the input Java methods to OpenCL C code. 


\paragraph{TornadoVM API:}
Since this work uses the TornadoVM API to execute the Deep Netts framework on heterogeneous devices, we present in this section an example of how TornadoVM is programmed. 
TornadoVM exposes an API that expresses the task-based parallelism, in which each task is a reference to an existing Java method. 
Additionally, the TornadoVM API can create a group of tasks that can be compiled together in the same \textit{compilation unit}, and subsequently run on the same target device (e.g., the same GPU).
This group of tasks is called a \texttt{TaskSchedule}.

Developers programming with TornadoVM can also annotate the code by using the \hcode{@Parallel} annotation that informs the TornadoVM JIT compiler that a loop is a candidate for parallel execution. 
Moreover, there is the \hcode{@Reduce} annotation that informs the compiler for a reduction operation, in which an input array is reduced to a scalar value computed with an associative and commutative operator.
The basic characteristic of the TornadoVM API is that it allows Java programmers to exploit hardware parallelism without requiring the knowledge of OpenCL or hardware architecture. 
As part of the fall-back execution feature, the annotated code is adapted based on the characteristics of the device, and in case of a single-threaded CPU execution it can be ignored.


\begin{figure}[t!]
\begin{lstlisting}[label=code:java, caption={Java code-snippet to run matrix multiplication by using TornadoVM.}, xleftmargin=.03\textwidth]
public class Compute {
    private static void mxm(Matrix2DFloat A, Matrix2DFloat B, Matrix2DFloat C) {
        for (@Parallel int i = 0; i < SIZE; i++) {
            for (@Parallel int j = 0; j < SIZE; j++) {
                float sum = 0.0f;
                for (int k = 0; k < SIZE; k++) 
                    sum += A.get(i, k) * B.get(k, j);
                C.set(i, j, sum);
            }
    }}
    public void run(Matrix2DFloat A, Matrix2DFloat B, Matrix2DFloat C) {
        TaskSchedule t = new TaskSchedule("s0")
             .task("t0", Compute::mxm, matrixA, matrixB, matrixC)
             .streamOut(matrixC);
             .execute();
	}}
\end{lstlisting}
\vspace{-2.5em}
\end{figure}

Listing~\ref{code:java} shows an example for computing a matrix multiplication by using the TornadoVM API. 
Lines 2-10 show the code of the method that encompasses the sequential implementation of the matrix multiplication. 
Lines 3 and 4 have been enhanced by using the \hcode{@Parallel} annotation in order to hint the TornadoVM compiler that these loops can be parallelized. 
Then, the method \texttt{run} (line 11) instantiates a \texttt{TaskSchedule} object with a single task, pointing to the \texttt{mxm} method. 
Finally, the code is executed in line 15. 
Note that the Java code is totally agnostic about the hardware device on which the program will be executed. 
Once the program is annotated, the TornadoVM compiler and runtime compile and execute all tasks enclosed within the \texttt{TaskSchedule}.

\section{How Deep Netts is parallelized with TornadoVM?}
\label{sec_implementation}

We extended Deep Netts to use TornadoVM for the training phase of the back-propagation stage for two particular layers: a \textit{fully connected} layer, and a \textit{softmax output} layer.
The two functions were modified in a similar way. 

\begin{figure}[!b]
	\centering
	\vspace{-1.5em}
	\includegraphics[scale=0.6]{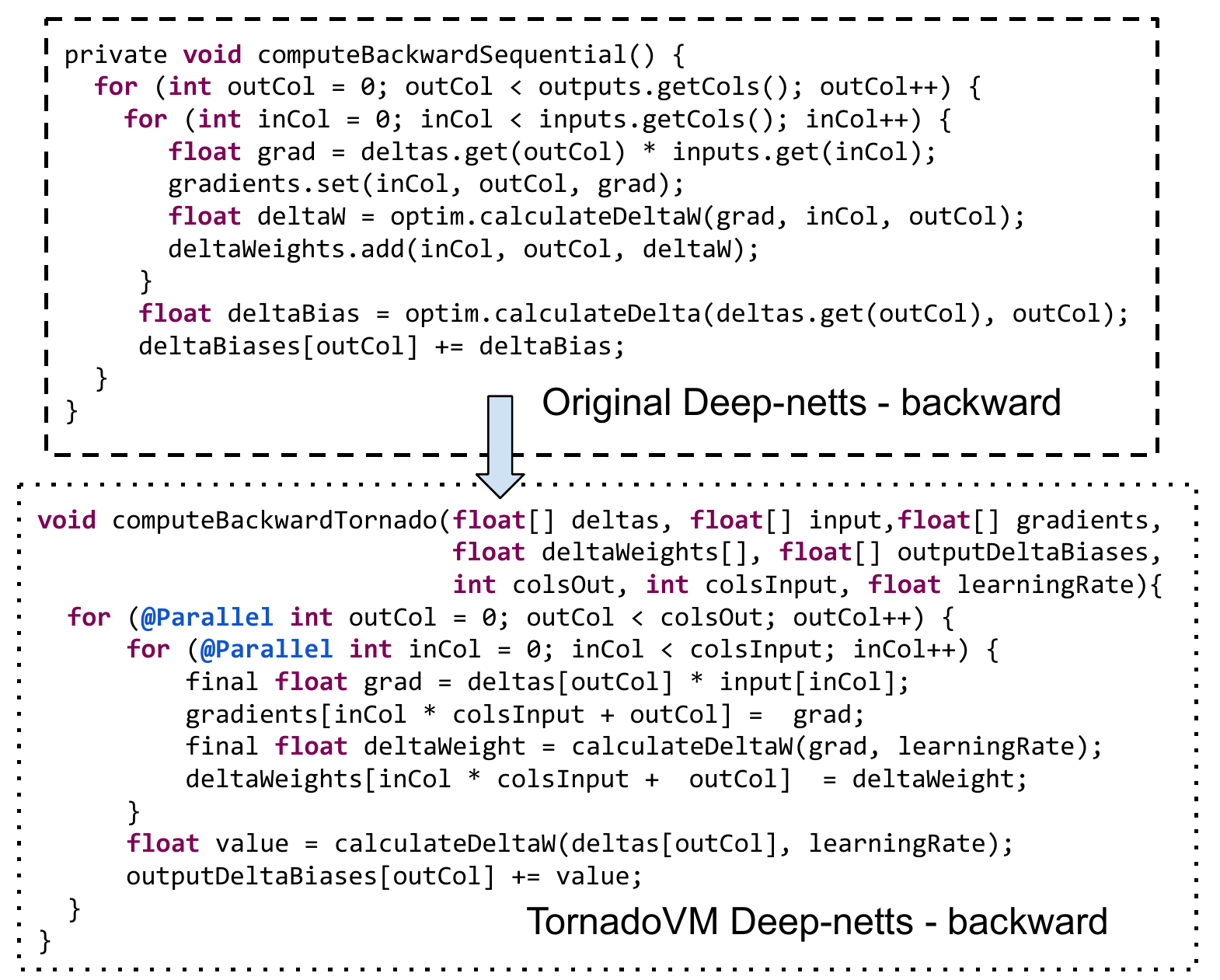}
	\caption{The softmax output layer code transformation that expresses the backward computation with TornadoVM.}
	\label{figure::soft}
	\vspace{-1.5em}
\end{figure}

In this paper we demonstrate the process of accelerating the method \texttt{backward} for one layer, the \textit{softmax output} layer due to space constraints. 
Figure~\ref{figure::soft} shows the transformations that were applied to the method \texttt{backward}.
The top part of Figure~\ref{figure::soft} shows the original version, while the bottom part shows the same Java snippet by using the TornadoVM.
At first, we needed to convert Java object types to primitive arrays. 
We used those primitive arrays to pass as method parameters to the TornadoVM's version. 
The second transformation was the usage of the \hcode{@Parallel} annotation, as discussed in Section~\ref{section::tornadovm}.
This annotation is used as a hint by the TornadoVM compiler to parallelize those loops with OpenCL. 
In this case, we annotated two loops, which potentially can use 2D parallel-indexing in the OpenCL side. 

Additionally, we needed to indicate to TornadoVM that we want to accelerate the annotated method. 
Thus, we instantiated the \texttt{TaskSchedule} object as shown in Listing~\ref{code:soft}.
We first declared the set of variables to copy into the target device and passed them as input to the \texttt{streamIn} method (line 2). 
Then we declared a task (line 3), which points to the method annotated with TornadoVM.
Finally we specified as input to the \texttt{streamOut} method (line 6) the variables that should be synchronized with the host (the main CPU) after the execution, and then we called the \texttt{execute} method (line 7). 

\begin{figure}[t!]
\begin{lstlisting}[label=code:soft, caption={TaskSchedule that builds the backward task from the SoftmaxOutputLayer class with TornadoVM.}, xleftmargin=0.03\textwidth]
TaskSchedule ts = new TaskSchedule("SoftwareOutputLayer");
ts.streamIn(tornadoDeltas, tornadoGradients, tornadoDeltaWeights, deltaBiases)
   .task("backward", SoftmaxOutputLayer::computeBackwardTornado, tornadoDeltas, 
            tornadoInputs, tornadoGradients, tornadoDeltaWeights, deltaBiases, 
            colsOut, colsInput, learningRate)
   .streamOut(tornadoGradients, tornadoDeltaWeights, deltaBiases)
   .execute();
\end{lstlisting}
\vspace{-2em}
\end{figure}

\subsection{Benefits of Our Approach}
The main benefits of our approach against the state-of-the-art deep learning frameworks are as follows:
\begin{enumerate}
	\item The heterogeneous execution of any part of the frameworks that is required by the user, while other works (e.g., TensorFlow~\cite{199317}) support precompiled binaries of particular functions.
	\item The ability to transparently compile and execute any part of the framework code gives the freedom to the developers to execute on any supported device such as a CPU, a GPU or an FPGA. On the contrary, other works (e.g., TensorFlow~\cite{199317}) are bounded to run various software versions on the specific GPU architectures that are supported.
	\item The execution of a deep learning framework that follows our approach can be migrated from a GPU to an FPGA, or even another GPU at runtime. This is a key feature for dealing with trade-offs regarding performance and power dissipation, or other runtime factors, such as device availability; that many of the existing frameworks lack of.
\end{enumerate}
\section{Experimental Evaluation}
To evaluate the Tornado-Deepnetts framework against the original Deepnetts (Original Deep Netts) implementation, we followed the following principles.
We used the same hardware device and the same software operating system and heap size, as presented in Table~\ref{label::platform_characteristics}.
We run the benchmark application as presented in Section~\ref{benchmarks}.
We performed the warm-up process which included 10000 executions prior to the actual timing of both systems, in order to provide fair results. The reported results in Section~\ref{performance_results} are the arithmetic average of 10 iterations of each measurement with respect to the JVM variance.

\begin{table}[h]
	\centering
	\vspace{-2.5em}
	\caption{The experimental hardware and software characteristics of our testbed.}
	\label{label::platform_characteristics}
	\begin{tabular}{|l|c|}
		\hline
		\textbf{CPU}  & Intel Core i7-8700K CPU @ 3.70GHz (Hyper-threading x12) \\ \hline
		\textbf{Memory} &  64 GB \\ \hline
		\textbf{JVM Heap Size} & 16 GB \\ \hline
		\textbf{GPU} &  AMD Radeon RX Vega 64 GPU \\ \hline
		\textbf{Operating System} & Ubuntu 18.04.01 (kernel 4.15.0-47-generic)   \\ \hline
		\textbf{Cmake} & v3.10.2  \\ \hline
		\textbf{Gcc} & v7.4.0  \\ \hline
		\textbf{Python} & v2.7.15  \\ \hline
	\end{tabular}
	\vspace{-2.5em}
\end{table}

\subsection{Benchmark Application}
\label{benchmarks}
In our experiment we evaluated the performance of the two backward methods presented in Section~\ref{sec_implementation} for the IrisClassification algorithm. For both systems we used the \textit{softmax} as the activation function in the output layer and the \textit{Cross Entropy} as the loss function. The size of the initial dataset that provided in the original Deep Nett framework is 6922 Bytes. 
At first, the IrisClassification algorithm loads the data set\footnote{The data set file is named ``iris\_data\_normalised.txt''.}. 
Then it uses the \texttt{dataSet.split()} method to split the data to 0.9 and 0.1. This means that 90\% of the data are used for training, and the remaining 10\% are used for testing. In the first step, the feed forward network is created and configured using the builder() method. 
The number of input features used in our experiment is 340, while the number of neurons in the \textit{fully connected} layer is 100000, and the number of possible categories in the output layer is 10. In addition the activation function of the \textit{fully connected} layer is configured to be \textit{Tanh}. 
Unlike the \textit{fully connected} layer, the activation function of the output layers is \textit{softmax}. Then the loss function of the algorithm is cross entropy. Afterwards, the algorithm creates and configures an instance of the back propagation trainer, which is the point that the two accelerated methods are called. 
At the end of the algorithm, the \texttt{neuralNet.train()} method is used for training, while the \texttt{neuralNet.test()} method is used for testing.

\subsection{Performance Analysis}
\label{performance_results}
\begin{figure}[t]
 \centering
 \begin{subfigure}[b]{0.45\textwidth}
     \centering
     \includegraphics[width=\textwidth]{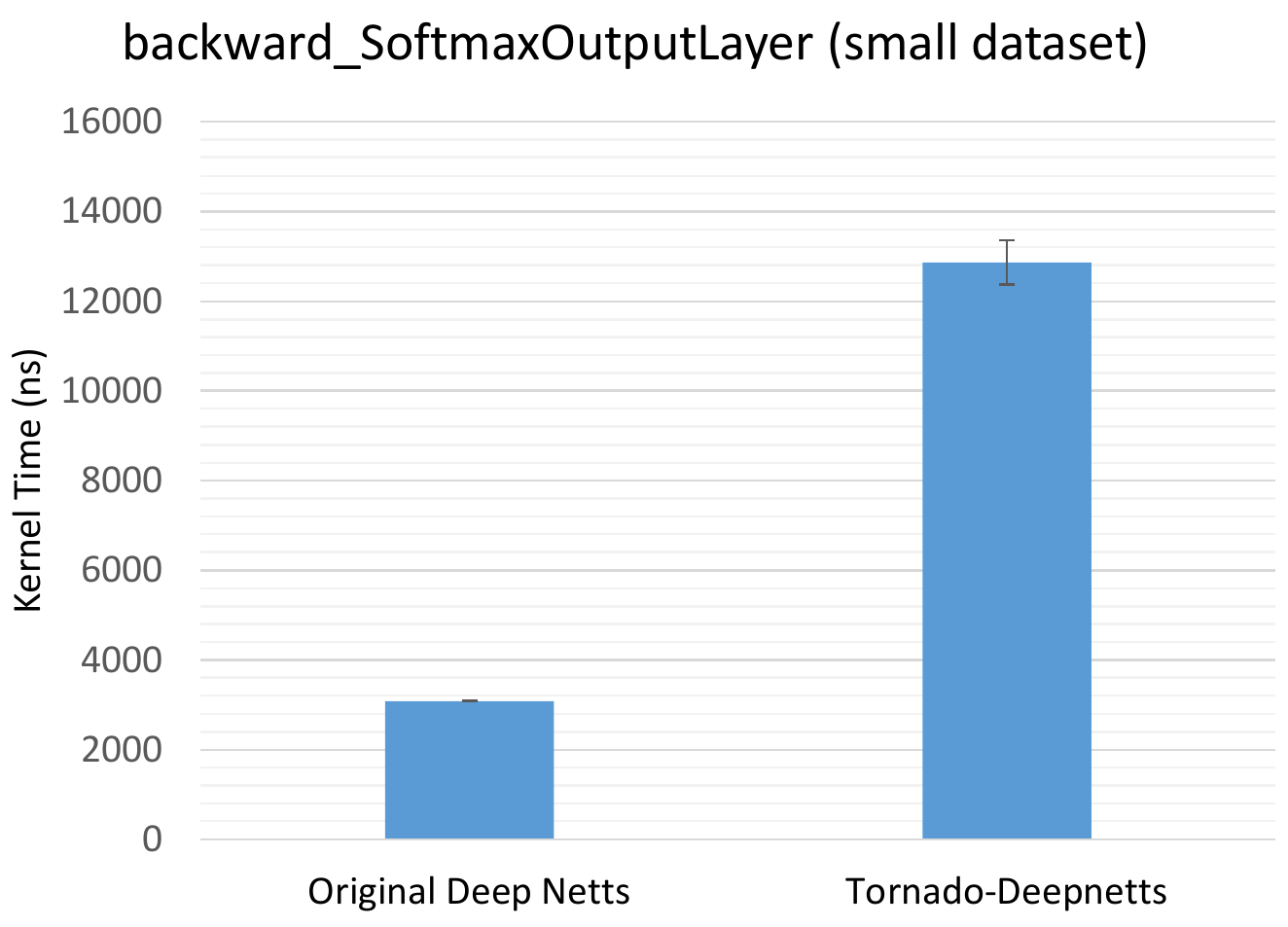}
     \caption{The softmax output performance.}
     \label{fig:softmax-small}
 \end{subfigure}
 \hfill
 \begin{subfigure}[b]{0.45\textwidth}
     \centering
     \includegraphics[width=\textwidth]{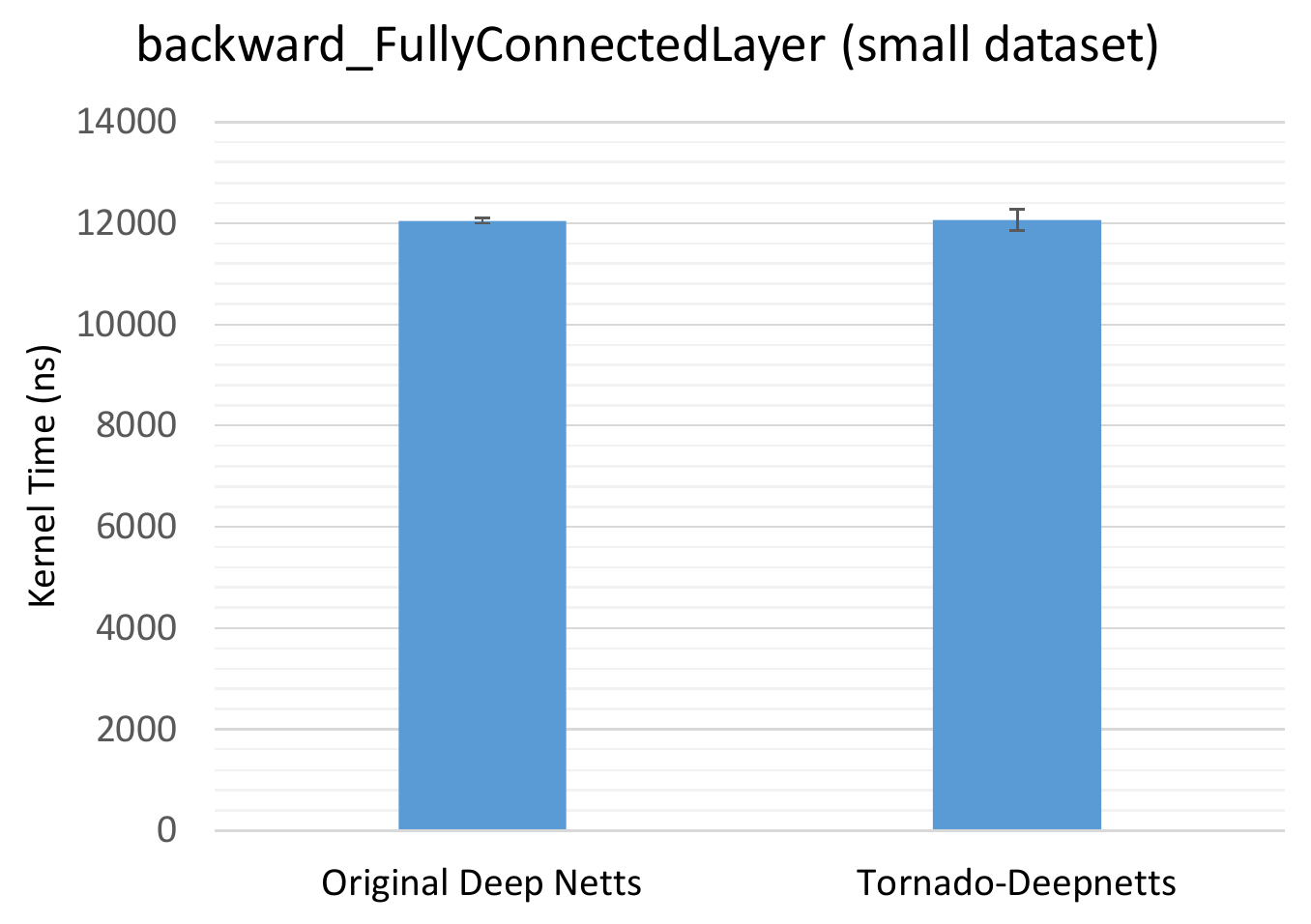}
     \caption{The fully connected performance.}
     \label{fig:fullyconnected-small}
 \end{subfigure}
    \caption{Comparison of two layers (i.e., (a) the softmax output layer and (b) the fully connected layer) in the backpropagation stage for small dataset, running through the Original Deep Netts and the Tornado-Deepnetts systems.}
	\vspace{-0.5em}
\end{figure}

The initial evaluation over the given dataset demonstrated that the computation was not sufficient to benefit from hardware acceleration. Figures~\ref{fig:softmax-small} and~\ref{fig:fullyconnected-small} show the performance on both evaluated systems (Original Deep Netts/Tornado-Deepnetts) for the dataset from Deep Netts. As shown in Figure~\ref{fig:softmax-small}, the backward method in the \textit{softmax output} layer performs up to four times slower on the GPU than the sequential JIT-compiled methods on the CPU, while the backward method in the \textit{fully connected} layer has almost identical performance on both systems. The main reason is that the dataset is small (7 Kilobytes) and the computation is not sufficient to overcome the cost of the data transfer from the host to the GPU and backwards.
  
\begin{figure}[t]
 \centering
 \begin{subfigure}[b]{0.45\textwidth}
     \centering
     \includegraphics[width=\textwidth]{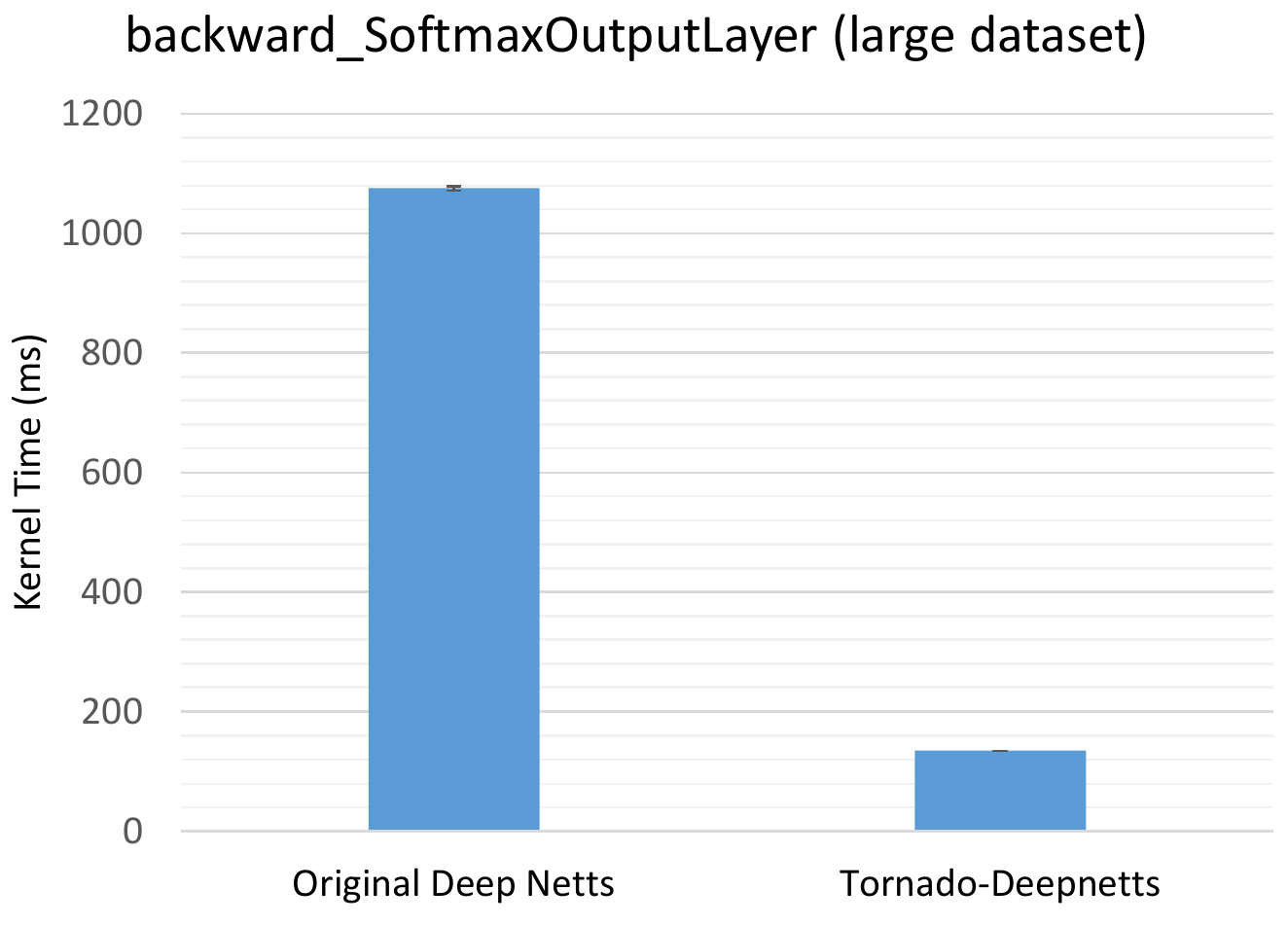}
     \caption{The softmax output performance.}
     \label{fig:softmax-large}
 \end{subfigure}
 \hfill
 \begin{subfigure}[b]{0.45\textwidth}
     \centering
     \includegraphics[width=\textwidth]{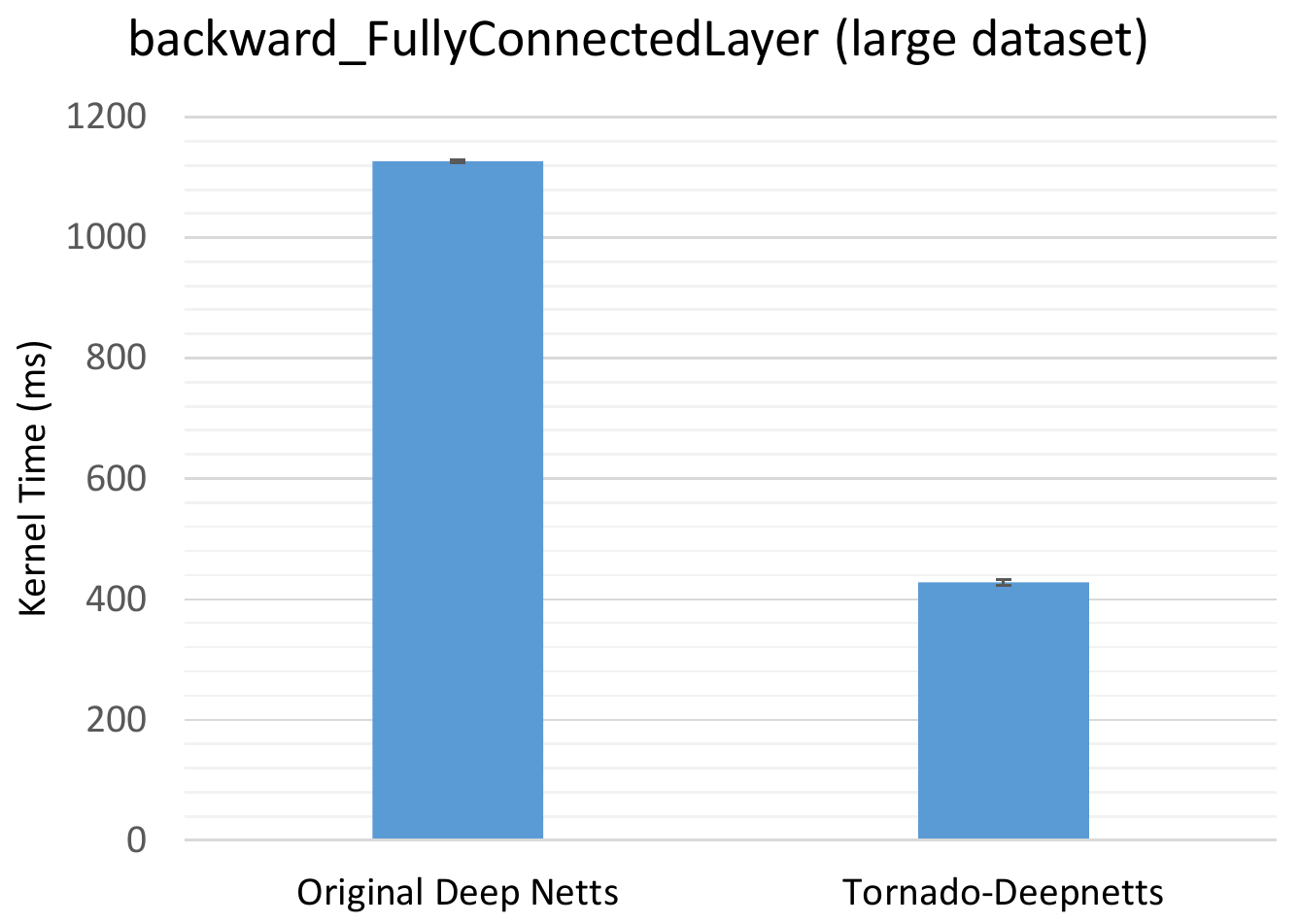}
     \caption{The fully connected performance.}
     \label{fig:fullyconnected-large}
 \end{subfigure}
    \caption{Comparison of two layers (i.e., (a) the softmax output layer and (b) the fully connected layer) in the backpropagation stage for large dataset, running through the Original Deep Netts and the Tornado-Deepnetts systems.}
	\vspace{-1.5em}
\end{figure}

To validate this case we populated the dataset randomly to increase the size  to 743424 Bytes, which is 107 times larger than the original dataset; and performed the same experiment a second time.  Figures~\ref{fig:softmax-large} and \ref{fig:fullyconnected-large} present the execution time of the two offloaded methods on both systems for the large dataset. Both times are reported in milliseconds. This time the Tornado-Deepnetts system outperforms the Original Deep Netts system, by  8x and 2.63x for the \textit{softmax output} layer and the \textit{fully connected} layer backward methods, respectively. In particular, the Tornado-Deepnetts reduces the execution time of the method from 1075 ms to 134 ms. Accordingly, the second method is reduced from 1126 ms to 427 ms.
\section{Summary}
In conclusion, this paper presents our work in progress towards accelerating deep learning engines written in Java on heterogeneous systems. 
Initially we studied Deep Netts, a deep learning engine fully implemented in Java that lacks of acceleration on heterogenous devices, such as GPUs and FPGAs. 
Then we employed TornadoVM, a state-of-the-art heterogeneous programming framework that enables software developers to target OpenCL-compatible devices for accelerating various Java workloads, without requiring any significant knowledge about hardware. 
Subsequently, we identified that the most expensive part of the framework occurred during the training of the model; and in particular during the backpropagation stage. 
The next step was the implementation of the Tornado-Deepnetts framework, by employing the TornadoVM API to accelerate the backpropagation stage in two of the available layers of the Deep Netts neural networks: the fully connected layer and the softmax output layer.
We evaluated the performance of the accelerated part in the implemented Tornado-Deepnetts framework against the original Deep Netts code. 
Our preliminary results showed that the selected methods can outperform the original Java methods by up to 2.63x and 8x for small and large datasets, respectively. 

As future work, we plan to: accelerate the remaining layers; conduct experiments with FPGA-based acceleration; share task schedules among the layer that could optimize any redundant copies of the current implementation; and investigate the Tornado-DeepNetts framework on distributed CPUs and GPUs.
\vspace*{-6pt}
\section*{Acknowledgements}
\vspace*{-7pt}
This work is partially supported by the EU Horizon 2020 E2Data 780245 grant.
We would like to thank Foivos Zakkak and Konstantinos Papangelou for their feedback.
\vspace*{-\baselineskip}

\bibliographystyle{splncs04}
\bibliography{sections/bibl}

\end{document}